# Grapheme-to-Phoneme Conversion using Multiple Unbounded Overlapping Chunks

François Yvon

École Nationale Supérieure des Télécommunications
Computer Science Department
46, rue Barrault 75 013 PARIS

**Abstract** We present in this paper an original extension of two data-driven algorithms for the transcription of a sequence of graphemes into the corresponding sequence of phonemes. In particular, our approach, a formal extension of the algorithm reported in [20], generalizes the algorithm originally proposed by Dedina and Nusbaum (D&N) [7], which had originally been promoted as a model of the human ability to pronounce unknown words by analogy to familiar lexical items. We will show that D&N's algorithm performs comparatively poorly when evaluated on a realistic test set, and that our extension allows us to improve substantially the performance of the analogy-based model. We will also suggest that both algorithms can be reformulated in a much more general framework, which allows us to anticipate other useful extensions. However, considering the inability to define in these models important notions like *lexical neighborhood*, we conclude that both approaches fail to offer a proper model of the analogical processes involved in reading aloud.

## 1 Introduction

Fast and accurate grapheme-to-phoneme conversion modules are an important requirement of many real-word applications, both in the domain of speech processing (synthesis and large-vocabulary speech recognition), and in the domain of natural language processing. For this task, most extant systems rely on a meticulously hand-shaped set of rewrite rules. This approach has proven to be definitely well-adapted to the task at hand, and various implementation have demonstrated its effectiveness [14]. However, it requires, for some languages (typically English or French), a very time-consuming handcrafting of large sets of rules. Moreover, this tedious task has to be restarted from scratch again every time a new language comes under focus. Finally, it seems that non-lexical items, typically proper names, are much too irregular to be transcribed by rules alone [4, 13].

As a consequence, a number of machine-learning techniques have been proposed as possible alternatives to rule-based systems for the transcription task: neural networks [17], decision-trees [9], instance-based learning [19], Markov models [2], analogy-based techniques [10], etc. The data-driven solution proposed by Dedina and Nusbaum [7] presents, with respect to the previous list, two main differences.

First, it relies, for the pronunciation of unknown words, upon large portions of existing transcriptions (hereafter referred to as *chunks*), when the other models induce and exploit a grapheme-per-grapheme classification mechanism. This has an important consequence, for it affects the nature of the object being studied. In a classification model, the central object is the grapheme[1], which is described by its graphemic neighbors, and has to be classified into the correct phonemic class; the lexicon of known words is consequently viewed as a mere stock of classified graphemes. In D&N's model, the important object is the chunk (i.e. a pairing between a graphemic and a phonemic substring of a known word), and the lexicon is primarily viewed as a stock of chunks. In fact, considering than many linguistic phenomena involved in the grapheme to phoneme conversion (tense vs lax alternations, stress shifts, etc) are expressed in terms of units which exceed the grapheme (morphemes, syllables, etc), we can already assert that this conception of the lexicon is probably more relevant from a linguistic standpoint than the one entertained in the letter classification approaches.

---
[1] In a very loose sense of the term: in most cases, the relevant object is indeed the letter.

Second, D&N's model is devised explicitly as a computer model for the human ability to form new pronunciations on the basis of previously acquired phonological representations, and therefore, it has cognitive implications. In fact, D&N explicitly refer to the experimental observations of Glushko [12], which offer some evidence regarding the effect of lexically similar items on the pronunciation of pseudo-words. Their model makes it possible to replicate this kind of effect, since it predicts correctly 91% of the observed human pronunciations of the pseudo-words used in Glushko's experiments.

As this per word accuracy outperforms by and large most extant data-driven systems, we reimplemented D&N's algorithm, and evaluated it on a more severe test-bed. We conducted various experiments (see section 4), which revealed that the generalization abilities of this algorithm are somewhat less satisfying than expected. A simple modification of their program, implemented in another algorithm (SMPA), nonetheless allows us to achieve much better results. With this new model, we will demonstrate the effectiveness of a non-classificatory approach to data-driven grapheme-to-phoneme conversion. We will also note that the our algorithm can be readily applied to various closely related tasks, such as phoneme-to-grapheme conversion, stress assignment, etc.

This paper is organized as follows: in section 2, we recall the context of D&N's work, and describe their algorithmic implementation. We then present carefully, in section 3, our own extension, and give in section 4 a number of experimental results. Section 5 is devoted to a discussion of both models, with respect to other chunks recombination techniques, and to the analogy model they implement. In particular, we conclude that SMPA, and the same holds for PRONOUNCE, despite its flexibility, its accuracy, and its ability to predict most of the observed human pronunciation of pseudo-words, cannot be considered as a proper model of pronunciation by analogy.

## 2 Dedina and Nusbaum's System

We give in this section a short summary of D&N's work. We will first recall the original background of their work, in relation to Glushko's model of reading-aloud, before presenting briefly the algorithmic aspects.

### 2.1 A Glance at the Experimental Results

The idea of a pronunciation system relying only on analogies originates from the experimental work of Glushko [12]. This author built a list of mono-syllabic pseudo-words, had them read aloud, and observed that:

– there were significant differences in the pronunciation latencies between two classes of pseudo-words (see above);

– a number of proposed pronounciations disagree with standard rules of English

These results do not fit in well with the traditional dual-route models of reading aloud (e.g. [5]), which assumes that pseudo-words are uniformely pronounced by abstract pronounciation rules. Under this assumption, none of the previous observations can be predicted.

As a typical example, the pseudo-word *tave* is read aloud both /tæv/ and /tev/, whereas *taze* is only pronounced (regularly) /tez/. In both cases, the standard pronunciation rules of English would predict a /e/. According to Glushko, the unexpected pronunciation /tæv/ is the result of the analogical pressure exerted by the irregular *have*, which conflicts with the regular *gave* in the lexical neighborhood of *tave*: this pseudo-word is thus said to have an inconsistent lexical neighborhood. Conversely, *taze* has only regular and consistent lexical neighbors (*gaze* and *daze*), and is therefore only pronounced in accordance with the rules. Glushko consequently questions the rule-based description, and suggests that a model of *lexical association* might be a better candidate to account for this kind of result. Additionally, such a model would also predict the differences in pronunciation latencies, a consistent lexical neighborhood facilitating the reading of an unknown string, as compared to pseudo-words having inconsistent neighbors.

However, it shall be noted that Glushko does not go much further in the specification of his model. In particular, he remains quite vague in his description of the lexical neighborhood, and simply outlines the priviledged role played by word endings (rhyme analogies) in the analogical processes. As an example of this underspecification of the lexical neighborhood, Glushko does not mention the existence of other lexical neighbors of *taze*, such as *tare* or *tape*, nor does he elucidate their potential influence.

## 2.2 The Original Algorithm

In accordance with Glushko's hypothesis, the basic idea of D&N's model is that an unknown word should be pronounced on the basis of lexical knowledge only (i.e. without any sort of rule-based mechanism). This is achieved by retrieving and recombining *large* chunks of existing lexical items. Given a dictionary of known words, the pronunciation of an unknown string $x$ proceeds according to the three following stages:

1. The first stage is to collect, in the list of examples, every string matching a substring of $x$. Each graphemic chunk is associated with a pronunciation, which is computed on the basis of an alignment between graphemic and phonemic sequences in the data-base;
2. The second stage consists in recombining these chunks so as to derive the complete pronunciation. This operation is achieved by concatenating, in all possible fashions, the transcriptions of the chunks extracted at stage 1. The only condition imposed on the concatenation operator is that two chunks can recombine if and only if the last phoneme of the first part is similar to the first phoneme of the second part. This constraint is implemented through the construction of a *pronunciation lattice* $\mathcal{L}(\S)$. Each traversal of this lattice corresponds to a possible transcription;
3. The final stage is to select, in the cases where multiple solutions are proposed, the best candidate. This selection is performed on the basis of the number of chunks involved in a solution: the lesser that number is, the better is the solution. This criterion reflects the intuition that pronunciation errors are more likely to appear at chunk boundaries than inside a chunk: intra-chunk graphemes are retrieved with large portions of their context, whereas graphemes occuring at chunk boundaries miss their right or their left context. As a consequence, minimizing the number of chunks involved in a solution tends to reduce *a priori* the chance of error.

Additionally, D&N use frequency data to break possible ties: when two pronunciations involve the same number of chunks, the one containing the most frequent ones is chosen. A last point is worth mentioning: given an unknown string $x$, and a reference lexicon, the existence of a transcription (i.e. a complete traversal of the pronunciation lattice) is not guaranteed. In other words, PRONOUNCE may not be able to pronounce some strings.

The graph on Figure 1 represents the pronunciation lattice corresponding to the word whose pronunciation is to be inferred, *hope*, given the following examples: *hot*,(/hɒt/), *hose* (/həʊz-/), *slope*, (/sləʊp-/), *slop* (/slɒp/), and *shop* (/ʃ-ɒp/).

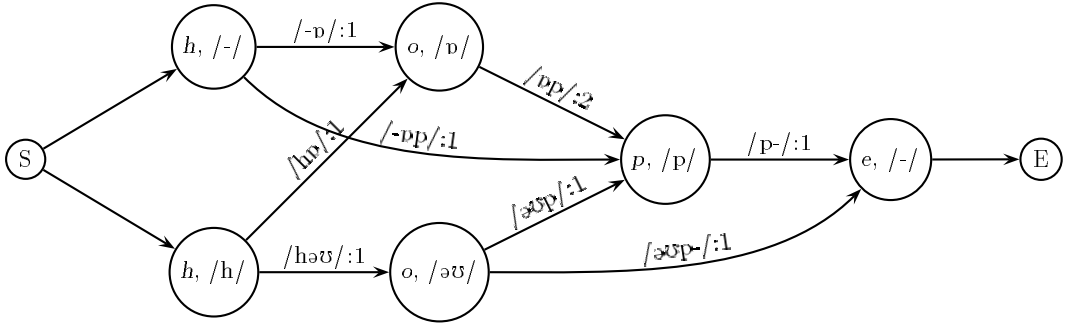

**Figure1.** The pronunciation lattice in PRONOUNCE

Each node in this graph accounts for the observation of a match with a grapheme in the unknown word: for example, the rightmost node, e, /-/, accounts for the match between a suffix of *hope* and a lexical item (*slope*). Two nodes are connected whenever a substring in the dataset contains the related sequence of phonemes: the rightmost node is thus connected to the node /p/, reflecting that the chunk *pe*, with the pronunciation /p-/, has been found (in *slope*). In this example, three different pronunciations can be inferred (each corresponding to a different path in the lattice): /həʊp-/, /-ɒp-/, and /hɒp-/. Using the shortest path criterion, we can select the first two pronunciations as the more reliable ones.

In order to introduce more clearly our extension to this algorithm, we can reformulate it as follows: PRONOUNCE transcribes unknown words by recombining existing phonemic chunks, on the condition that two adjacent parts *share exactly one common phoneme*.

## 3 Transcription with Unbounded Overlap: SMPA

### 3.1 Recombining Two Overlapping Chunks

In a previous experiment [20], we reported that a quite accurate transcription system could be based on a very simple principle, involving the recombination of the tail and a head of words whose spelling and pronunciation are known. As an example of this strategy, consider how the unknown word *énergence* can be pronounced by using chunks from two familiar items, *énergie* and *emergence*. This process is graphically pictured on Figure 2.

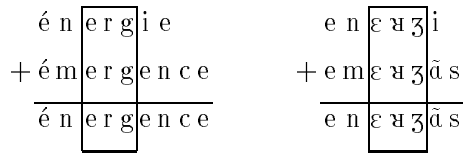

**Figure 2.** The head'n tail strategy

As this graphical representation suggests, the success of the head and tail strategy depends heavily on the size of a common portion in the head and in the tail part, both at the graphemic and at the phonemic level. This common portion, which appears in a framed box on Figure 2, will be refered to as the *overlap*. A large overlap indeed reduces the chance of undesirable contextual effects at the the junction of the two chunks. Conversely, transcriptions which are produced on the basis of small overlaps will tend to be far less reliable [11].

A major problem with this over-simple approach is that a significant number of unknown words might not be decomposable into an existing head and tail. For example, in a series of 10 experiments with the Nettalk lexicon [17], each involving an 18 000 word learning set and an independent 2 000 word test set, the percentage of words that could not be pronounced averaged 15%.

### 3.2  Recombining $N$ Overlapping Chunks

The natural extension of the head and tail strategy is to consider, as D&N do, the recombination of any number of chunks, while retaining the idea of favoring transcriptions which are based on *large overlapping chunks* (when their model only considers overlappings of length 1).

Algorithmically, the generalization of D&N's model to unbounded overlappings comes quite naturally. The only necessary modification of their program is to modify the definition of the pronunciation lattice $\mathcal{L}(\S)$ of an unknown word $x$ in the following manner:

- the nodes of $\mathcal{L}(\S)$ correspond to the phonemic chunks associated in the lexicon with a substring of $x$;
- the arcs of $\mathcal{L}(\S)$ join two nodes if and only if the related strings strictly overlap (i.e. one is not included in the other) both at the phonemic and graphemic level. These arcs are weighted on the basis of the overlap size;
- $\mathcal{L}(\S)$ has two additional vertices $S$ and $E$, which have an outgoing arc to any chunk matching respectively a prefix or a suffix of $x$. The weight of these arcs is not taken into account.

The design of this pronunciation lattice is such that every path from $S$ to $E$ corresponds to a possible transcription. Again, such a path does not necessarily exist: in that case, SMPA does not generate any pronunciation.

The graph reproduced on Figure 3 represents the pronunciation lattice for the unknown word *hope*; the lexicon is the same as the one we used in section 2.2.

The three paths from $S$ to $E$ on Figure 3 correspond to three possible pronunciations: /hɒp-/, /həʊp-/, and /-ɒp-/.

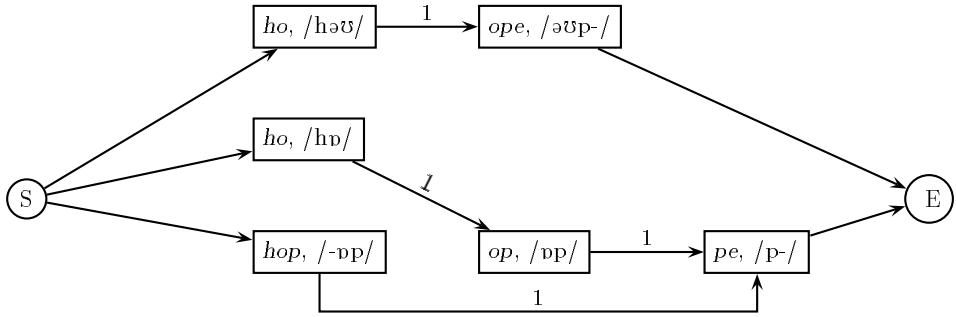

**Figure 3.** The pronunciation lattice in SMPA

We now have in hand a mechanism for producing transcriptions of an unknown graphemic string. The next step is to define an order on the set of the possible alternative pronunciations. As we wish to favor the solutions relying on large overlaps, each pronunciation is given a score, according to the following formula. If $\pi(x)$ is a possible pronunciation of an unknown word $x$, produced by following the path $\mathcal{P}$, where $|\mathcal{P}|$ nodes, corresponding to the chunks $\{s\}$, are visited, then its score is given by ($l(z)$ denotes the length of a string $z$):

$$C(\mathcal{P}) = \frac{\sum_{s \in \mathcal{P}} l(s)}{|\mathcal{P}| \times \updownarrow(\S)} \tag{1}$$

The intuition behind the particular choice of this scoring function can be expressed as follows: whenever a word $x$ is entirely found in the lexicon, there will be a direct path from $E$ to $S$ in the pronunciation graph, which visits only the node corresponding to the entire word. In that case, the score value is 1 ($l(x)/l(x)$). For an unknown word, the best we can expect to find is two chunks that overlap maximally (i.e. that have all but one character in common). These two chunks will necessarily have a length equal to: $l(x) - 1$, hence the score of the related pronunciation: $(l(x) - 1)/l(x)$. If such a solution cannot be found, the next best configuration is to use one chunk of length $l(x) - 1$ and one of length $l(x) - 2$, etc. At one point, however, it might be preferable to consider solutions which rely upon three or more chunks: this is the case when recombining just two chunks cannot be done with a reliable enough (i.e. large enough) overlap. The order our scoring function defines on the set of possible solutions precisely performs this kind of compromise between the number of chunks involved, and the total size of the overlap. Algorithmically, finding the best pronunciation according to that score corresponds to finding the path of minimum weight in an acyclic graph ($\mathcal{L}(\S)$).

Following D&N, we use frequency data to break the possible ties: when two solutions have the same score, the one that is supported by the more frequent chunks is favored. In the example pictured on Figure 3, two solutions (/həʊp/ and /-ɒp-/) have the same score (0.625). All the chunks involved having the same frequency, this additional criterion cannot, in that oversimple case, break the ties.

The complexity of this algorithm can be analyzed as follows:

- the matching process requires every substring of length greater than 2 in $x$ to be searched in the lexicon, corresponding at worst to $l(x).(l(x) - 1)/2$ searches (in our implementation, all the chunks available in the lexicon are stored in a finite-state automata, making each search linear in the length of the substring, that is, at worst $l(x)$);
- the transcription process itself requires a minimal weight path in the acyclic graph $L(x)$ to be found. This procedure is linear in the number of vertices $V(x)$. Assuming that there is a upper bound on the maximum number of pronunciations associated with a graphemic chunk, say $K$, we can derive an upper bound for the number of nodes $N(x)$ in $L(x)$: $K \times l(x) \times (l(x)+1)/2$. The maximum number of vertices being bounded by $N(x)^2$, the complexity of the recombination stage is thus $O(l(x)^4)$;

To sum up, the complexity of the algorithm is $O(l(x)^4)$, which makes it, given the typical size of $l(x)$, computationally very tractable.

## 4  A Comparative Evaluation

The only evaluation reported by D&N uses 70 pseudo-words designed by Glushko. These pseudo-words were built by changing the initial consonant in existing mono-syllabic words. It is therefore rather unsurprising, considering how their algorithm proceeds, that they could accurately simulate and replicate Glushko's observations.

We have reimplemented PRONOUNCE[2], and have tested it on more realistic data. For this experiment, we have used a number of lexicons from different languages. Two of them are public-domain pronunciation dictionaries: NETTALK [17], an English lexicon, and BDLEX [15], a French lexicon. Additionally, we used three proper-name lexicons[3] created during the ONOMASTICA [16] project. All these lexicons contained between 20,000 and 25,000 examples.

For each lexicon, we have used the same experimental design: 10 pairs of disjoint (learning set, test set) were randomly selected and evaluated. In each experiment, the test set contains about the tenth of the available data. A transcription is judged to be correct when it matches exactly the pronunciation listed in the database at the segmental level. The number of correct phonemes in a transcription is computed on the basis of the string-to-string distance with the target pronunciation. For both algorithms, the proportion of words which could not be pronounced at all was reasonable (ranging from about 0.5% on NETTALK to about 2.5% on BDLEX). A summary of the performance is shown in Table 1.

| Corpus | PRONOUNCE | | SMPA | | DEC | |
|---|---|---|---|---|---|---|
| | % words | % phonemes | % words | % phonemes | % words | % phonemes |
| NETTALK | 56.56 | 91.69 | 63.96 | 93.19 | 56.67 | 92.21 |
| BDLEX | 82.97 | 96.65 | 86.54 | 95.43 | 86.45 | 97.86 |
| IT-NP | 93.07 | 98.82 | 95.73 | 99.06 | 96.02 | 99.43 |
| FR-NP | 74.25 | 94.38 | 79.14 | 95.62 | 73.63 | 95.08 |
| NE-NP | 81.66 | 96.59 | 89.86 | 98.57 | 89.16 | 97.95 |

Table 1. Comparative evaluation of PRONOUNCE, SMPA and DEC

---

[2] Allthough we have followed quite precisely the implementation details given in [7], we were not able to reproduce their results. Taking NETTALK as the reference lexicon, our own implementation achieves about 72% per word accuracy on Glushko's pseudo-words. Similar difficulties in reproducing D&N's results have also be reported by Damper [6].

[3] I would like to express my gratitude to Vito Pirrelli from the ILC (Pisa), and to Henk ven Heuvel from the KUN (Nijmegen), for providing these respectively Italian (IT-NP) and Dutch (NE-NP) corpora. The production of the French corpus has been done in our group.

As can be seen, on these lexicons[4], SMPA systematically outperforms PRONOUNCE, all the differences being statistically significant at a 0.01 confidence level in a two-tailed test of mean comparison on paired samples.

Additional comparison sources are provided by the evaluations of DEC [18], a decision-tree learning algorithm. We have also reimplemented this algorithm, and performed an evaluation on the same data bases. The decision-tree approach enables to achieve accuracy scores that are in-between those of PRONOUNCE and SMPA: very close to PRONOUNCE's performance on NETTALK and on FR-NP, DEC's results ressemble SMPA's on IT-NP, NE-NP and BDLEX. Still, SMPA outperforms DEC in a test of mean comparison, the differences in terms of word accuracy being statistically significant (at the 0.01 confidence level) for all lexicons. Note however that, as far as the per phoneme accuracy is concerned, DEC outperforms SMPA on two lexicons: BDLEX and NP-IT. A careful study of the transcription errors reveals that in many cases, when SMPA cannot produce a transcription (i.e. gets all phonemes wrong), DEC outputs an incorrect one. If we compute SMPA's performance on the basis of the number of transcriptions it produces, that is, if we do not take the silence into account, then the superority of the chunk-based strategy is statistically proven, even in terms of the percentage of correct phonemes.

If our assumption that DEC represents the state-of-art of classificatory pronunciations systems[5] is correct, then these results seem to give some insight regarding the fruitfulness of a paradigm shift from classification-based pronunciation to chunk-recombination based pronunciation.

## 5 Perspectives and Discussion

### 5.1 Perspectives

A number of improvements to the algorithm hereabove presented are currently under investigation. For example, we are investigating the effectiveness of using more "natural" alignment schemes. At the moment, we use one-to-one alignments of the graphemic and the phonemic strings to derive the pairings between graphemic and phonemic chunks. In our context, it seems that the use of many-to-many alignment schemes might be an efficient way to get rid of many unreliable chunks. This is obvious when we look at a word like *chanson*, and its transcription /ʃ-ã-sɔ̃-/, whose longest chunks are listed in Table 2.

| chanso-/ʃ-ã-sɔ̃/ | hanson-/-ãsɔ̃-/ |
|---|---|
| chans-/ʃ-ã-s/ | hanso-/-ãsɔ̃/ |
| anson-/ã-sɔ̃-/ | chan-/ʃ-ã-/ |
| hans-/-ã-s/ | anso-/ã-sɔ̃/ |
| nson-/-sɔ̃-/ | cha-/ʃ-ã/ |

**Table 2.** Some chunks extracted from the lexical item *chanson*

---

[4] Interestingly, SMPA has also been tested on Glushko's 70 pseudo-words, using NETTALK as a training set, and achieved, on this list, a nice 84.5% per word accuracy.
[5] Better performance can arguably be obtained using classification techniques (e.g. [1]), but still, as far as we know, significantly higher per word accuracy scores have always been observed in experimental conditions involving the use of some kind of additional learning bias.

Chunks like (ch*a*, /ʃ-ã/) are indeed very unreliable, since this correspondence is only observed in the context of a following *n*. These chunks could be readily discarded using a more adapted lexical representation of this word, where the digraph *an* would be aligned with the sound /ã/. This should improve the performance of the transcription mechanism both in terms of the resources it requires (a smaller amount of chunks will need to be stored), and in terms of the transcription accuracy.

From a more general standpoint, we believe that the following reformulation of both models in a more general framework might also be extremely profitable. Given a finite lexicon, we can extract a finite number of chunks, each of which can be considered as a particular symbol of a finite alphabet $C$. As SMPA transcribes any word which decomposes into a sequence of overlapping chunks, the mechanism it implements is able to parse any sequence of symbols of $C$, under the condition that two adjacent symbols correspond to overlapping chunks. This condition restricts the acceptable sequences in $C$; in fact, it defines a particular language $\mathcal{L}$ on the strings of chunks. This language is obviously regular, as can be readily observed by the pronunciation lattice of Figure 3 (see also [21] for a more detailled account). This lattice represents indeed a small portion of the finite-state automata corresponding to $\mathcal{L}$.

The transcription mechanism implemented in SMPA and in PRONOUNCE can therefore be considered as a particular instantiation of a regular grammar on an alphabet of chunks. Multiple parses in this highly ambiguous grammar are ranked according to the structural criterion defined in equation 1. Within this framework, a number of extensions can be considered. As an example, this grammar could be stochastized. What remains to be seen is whether a stochastic version of SMPA could simultaneously retain the idea of a structural ranking over the set of possible parses, while providing a more principled framework for using frequency data. Results obtained using a quite related model [8] (where chunks are not bound to overlapping conditions) [8] seem to indicate that, within a probabilistic framework, the use of large chunks is indeed implicitly favored. This seems to be a direct consequence of the computation of the likelihood of a word, which decomposes into the product of the probabilities of the chunks involved in the transduction: the smaller the number of chunks, the smaller the number of terms in the product, and the higher the likelihood.

## 5.2 Discussion

In this paper, we have proposed and evaluated an extension of D&N's algorithm. Our transcription mechanism demonstrates, more clearly than PRONOUNCE does, the effectiveness of a non-classificatory approach to the grapheme-to-phoneme conversion task. Furthermore, we have suggested that both algorithms belong to a much more general class of models, where the learning of a transduction mechanism is performed through the instantiation of a regular grammar defined on the set of all available chunks. This reformulation allows us to anticipate further fruitful developments for these techniques.

In our view, however, these rather encouraging result have little to say regarding the relevance of chunk-based algorithms as models of the human ability to read aloud unknown words. In fact, the objects which are really of interest for SMPA and PRONOUNCE are the so-called chunks, i.e. the portions of existing lexical items that can be used in a recombination process. Given an unknown input whose pronunciation is to be induced, the number of chunks involved in the output transcription will necessarily be restricted: this is a consequence of the particular function used to rank alternative recombinations. This, however, does not give any clue regarding the amount of different lexical items which eventually contribute to this transcription. In other words, it seems difficult to define in a chunk-based model the notion of "analogous lexical item", and, consequently, of "lexical neighborhood". In fact, considering that any word that is involved in the pronunciation of the unknown word is an analogue would lead to very unintuitive analogies, but these models do not offer any alternative ways to define these concepts.

Still, a proper cognitive model of pronunciation by analogy should contain a clear definition of these terms, in order to simulate or test a number of effects that have been reported or hypothesized, such as the primacy of word endings in analogies, the differential analogical pressure exerted by words of varying frequencies, the role of phonological analogies, etc [3].

Additionally, from a strict computational standpoint, it might be very profitable to take advantage of richer lexicons for the grapheme-to-phoneme conversion, i.e. lexicons storing more than just the graphemic and phonemic character strings. We have so far implicitly considered that a lexicon should contain only graphemic and phonemic information, but it is well-known that enriching pronunciation dictionaries, for example with part-of-speech tags, is the only means to make a transcription system predict correctly the pronunciation of homographous heterophons. One might wish to go further, and propose a model where a "verb-like" unknown word would be primarily analogized with other verbs, etc. This implies that the representation of the lexicon be able to accomodate this kind of supplemenatary information, and that the transcription model be able to use profitably such data. Again, it is very unclear how this could be done in a chunk-based model.

We shall therefore conclude that, despite their ability to replicate most of the pronuncations observed by Glushko, neither SMPA, nor PRONOUNCE, provide appropriate models of analogy-based processing for the task of reading aloud.